\documentclass[runningheads]{llncs}
\usepackage{graphicx}
\usepackage{cite}
\usepackage{amsmath,amssymb,amsfonts}
\usepackage{graphicx}
\usepackage{textcomp}
\usepackage{xcolor}
\usepackage{bm}
\usepackage{amsmath}
\usepackage{amssymb}

\usepackage{amsthm}
\usepackage{mathrsfs}
\usepackage{enumerate}
\usepackage{multirow}
\usepackage{color}
\usepackage{threeparttable}
\usepackage{adjustbox}

\usepackage[square, comma, sort&compress, numbers]{natbib}
\usepackage[pagebackref=false,breaklinks=true,letterpaper=true,colorlinks,bookmarks=false]{hyperref}
\usepackage{times}
\usepackage{breakurl}
\usepackage{array}
\usepackage{verbatim}
\usepackage{algorithm}
\usepackage{bbding}
\usepackage{algpseudocode}
\usepackage{lettrine}
\usepackage{float}
\usepackage{caption}
\usepackage{subcaption}
\usepackage{pifont}
\def\BibTeX{{\rm B\kern-.05em{\sc i\kern-.025em b}\kern-.08em
    T\kern-.1667em\lower.7ex\hbox{E}\kern-.125emX}}
\begin{document}
\title{YOLO-FDA: Integrating Hierarchical Attention and Detail Enhancement for Surface Defect Detection}
%
%
\author{Jiawei Hu\inst{1}\orcidID{0009-0007-2476-3124}}
\authorrunning{J. Hu}
%
\institute{College of Mechanical and Electrical Engineering, China Jiliang University, Hangzhou, Zhejiang, China\\
  \email{2200110409@cjlu.edu.cn}%
  }
\maketitle              
\begin{abstract}
Surface defect detection in industrial scenarios is both crucial and technically demanding due to the wide variability in defect types, irregular shapes and sizes, fine-grained requirements, and complex material textures. Although recent advances in AI-based detectors have improved performance, existing methods often suffer from redundant features, limited detail sensitivity, and weak robustness under multiscale conditions.
To address these challenges, we propose YOLO-FDA, a novel YOLO-based detection framework that integrates fine-grained detail enhancement and attention-guided feature fusion. Specifically, we adopt a BiFPN-style architecture to strengthen bidirectional multilevel feature aggregation within the YOLOv5 backbone. To better capture fine structural changes, we introduce a {Detail-directional Fusion Module (DDFM)} that introduces a directional asymmetric convolution in the second-lowest layer to enrich spatial details and fuses the second-lowest layer with low-level features to enhance semantic consistency. Furthermore, we propose two novel attention-based fusion strategies, {Attention-weighted Concatenation (AC)} and {Cross-layer Attention Fusion (CAF)} to improve contextual representation and reduce feature noise.
Extensive experiments on benchmark datasets demonstrate that YOLO-FDA consistently outperforms existing state-of-the-art methods in terms of both accuracy and robustness across diverse types of defects and scales.

\keywords{BiFPN  \and Detail-directional Fusion Module \and Attention-weighted Concatenation \and Cross-layer Attention Fusion.}
\end{abstract}

\begin{figure}[t]
    \centering
    \includegraphics[width=1\linewidth]{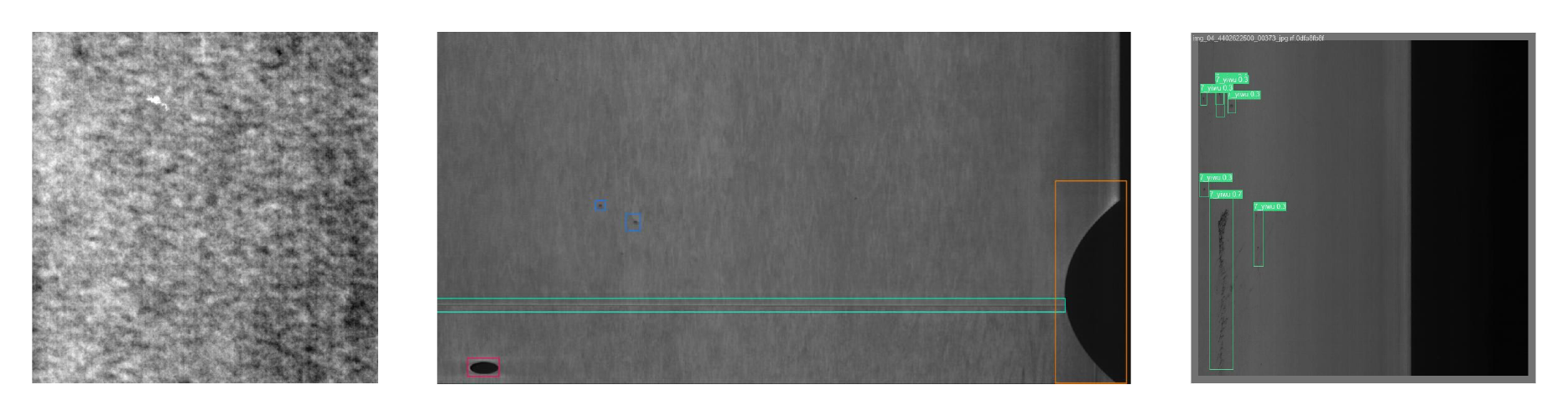}
\caption{Examples of surface defect from DAGM2007 and GC10-DET.}
\vspace{-1cm}
\label{fig:examples of surface defect from DAGM2007}
\end{figure}

\section{Introduction}
Surface defects in industrial manufacturing often compromise the structural integrity of products and may lead to significant economic losses. Traditional manual inspection methods are labor-intensive, error-prone, and highly susceptible to environmental and human factors, making it difficult to ensure consistent quality assurance. With the rapid development of deep learning, intelligent defect detection methods have attracted increasing attention due to their high accuracy, scalability, and robustness. In particular, real-time detection powered by AI has emerged as a promising solution for preventing production faults at the source~\cite{shen2025long}. However, challenges persist in detecting small-scale, irregularly shaped defects under complex background conditions. These difficulties largely stem from insufficient detail preservation, suboptimal multi-scale feature fusion, and limited contextual awareness in current detection frameworks~\cite{shen2025imagharmony}.

Early object detection frameworks, such as the two-stage R-CNN series~\cite{girshick2014rich,7410526,Shaoqing,he2017mask,cai2018cascade}, demonstrated strong localization capabilities but suffered from high computational costs and long inference times due to their reliance on explicit region proposals. While subsequent works such as Guided Anchoring~\cite{wang2019region} and Cascade RPN~\cite{vu2019cascade} optimized region proposal mechanisms, the two-stage pipeline remains inefficient for real-time industrial deployment.
To address these limitations, one-stage detectors such as SSD~\cite{liu2016ssd} and the YOLO series have emerged as more practical alternatives, offering faster inference and simplified training. YOLOv3~\cite{redmon2018yolov3} introduced Feature Pyramid Networks (FPN) to enhance multi-scale fusion, while YOLOv4~\cite{bochkovskiy2020yolov4} further improved the fusion architecture by incorporating PANet. More recently, YOLOv8~\cite{10473783} adopted an anchor-free approach and modified the neck structure to improve detection speed and generalization. These advances have been extended to surface defect detection through customized lightweight modules and enhanced feature extractors~\cite{liuMetalSurfaceDefect2025,du2024aff,yasir2023multi,lu2024yolo}. Similar to the trend in virtual try-on and human-centric generation tasks~\cite{shen2025imaggarment,shen2025imagdressing}, where specialized structures are designed to capture fine-grained textures and spatial semantics, industrial defect detection also demands careful modeling of both local details and global layout priors.

Despite these improvements, several limitations remain when applying YOLO-based models to surface defect detection. First, as shown in the first picture in Fig.\ref{fig:examples of surface defect from DAGM2007}, the diversity of defect types across materials such as steel, glass, and fabric, along with subtle variations in texture, gloss, and reflectivity, often leads to high feature noise and low discriminability in complex industrial backgrounds~\cite{shen2025imagharmony}. Second, many defects are extremely small or elongated, causing them to be lost during downsampling and poorly represented by conventional feature fusion strategies~\cite{shen2024imagpose}. As shown in the blue box in the second picture in Fig.\ref{fig:examples of surface defect from DAGM2007}, the defect category called inclusion (In) in GC10-DET usually appears in a small form, while the defect category called weld in GC10-DET, shown in the green box, usually has a larger aspect ratio. Third, as shown in the last picture in Fig.\ref{fig:examples of surface defect from DAGM2007}, while recent YOLO variants attempt to incorporate complex fusion modules, they often introduce information redundancy, which may degrade detection performance without benefiting semantic precision or spatial consistency~\cite{shen2025long}.

To address the aforementioned challenges in surface defect detection, we propose a novel framework named {YOLO-FDA}, which enhances both detail extraction and feature fusion in YOLO-based architectures. Specifically, we replace the original PANet neck of YOLOv5 with a more powerful {BiFPN} structure to reinforce bidirectional multi-scale information flow. To better preserve and emphasize fine-grained features of small or irregular defects, we design a {Detail-directional Fusion Module (DDFM)} to capture directional textures and explicitly merge low-level spatial details with higher-level semantics. 
In addition, we introduce two novel attention-based fusion mechanisms: {Attention-weighted Concatenation (AC)} ensures feature richness by preserving all input channels while assigning adaptive weights, and {Cross-layer Attention Fusion (CAF)} performs selective channel-wise weighting to suppress redundant or irrelevant features. These modules collectively improve the model’s ability to detect subtle defects in cluttered backgrounds.

The main contributions of this paper are summarized as follows:
\begin{itemize}
    \item[\textcolor{black}{$\bullet$}] We replace the PANet module in YOLOv5 with a BiFPN structure to strengthen bidirectional multi-scale feature fusion. Then, we introduce a {Detail-directional Fusion Module (DDFM)} to enhance fine-grained feature representations and improve sensitivity to localized defects.
    \item[\textcolor{black}{$\bullet$}] We propose two attention-based fusion strategies: {Attention-weighted Concatenation (AC)}, which preserves feature richness during fusion, and {Cross-layer Attention Fusion (CAF)}, which performs channel-aware weighting to reduce feature redundancy and improve interpretability.
    \item[\textcolor{black}{$\bullet$}] Experimental evaluations on the GC10-DET and DAGM2007 datasets demonstrate that YOLO-FDA achieves superior performance compared to state-of-the-art surface defect detection methods in terms of accuracy and robustness.
\end{itemize}

\section{Related Work}
\subsection{YOLO-Based Detection for Surface Defects}
The YOLO (You Only Look Once) family has rapidly evolved from YOLOv1~\cite{redmon2016you}, which enabled fast end-to-end detection, to modern variants like YOLOv5~\cite{1234yolov5} and YOLOv8~\cite{10473783}, which incorporate CSPDarknet, PANet, and anchor-free designs to improve accuracy and efficiency. Yolo-World~\cite{cheng2024yolo} further expands YOLO’s applicability in open scenarios through visual-language pretraining.
For surface defect detection, SLF-YOLO~\cite{liuMetalSurfaceDefect2025} leverages channel gating and lightweight necks for efficiency, while QCF-YOLO~\cite{zhouQCFYOLOLightweightModel2025} enhances tiny defect detection using GhostConv and a P2-head. However, these models still struggle with complex backgrounds and irregularly shaped defects due to limited multi-scale fusion and redundant feature aggregation.
To address these issues, we propose incorporating asymmetric convolution for detail modeling and cross-layer attention fusion for redundancy suppression—drawing inspiration from frequency-domain interaction and layout-aware enhancements in aerial and SAR detection~\cite{weng2023novel,qiao2022novel}.

\subsection{Attention Mechanisms in Defect Detection}
The introduction of Transformer~\cite{vaswani2017attention} sparked widespread adoption of attention mechanisms in visual tasks. Modules like Squeeze-and-Excitation (SE)~\cite{hu2018squeeze} and CBAM~\cite{woo2018cbam} have been widely used to adaptively recalibrate channel and spatial features. These attention designs have also been applied in defect detection, such as MCBAM used in YOLO-HMC~\cite{10384383}, which effectively enhances detection of small targets.
However, traditional attention modules remain limited. SE operates only within individual channels, making cross-scale feature comparison difficult, while CBAM applies spatial and channel attention serially without joint optimization. Inspired by prior work in selective frequency modeling~\cite{weng2024enhancing} and refined pyramid architectures~\cite{li2024lr}, we enhance attention integration via two novel modules: AC, which preserves rich features without information loss; and CAF, which performs weighted merging while retaining the original channel dimension, thus reducing information redundancy. These attention strategies are tailored to maintain semantic consistency across scales, an approach consistent with frequency-domain insights in recent remote sensing research~\cite{weng2023novel}.

\section{Proposed Method}
\subsection{Overview}

\begin{figure*}[t]
    \centering
    \includegraphics[width=1\linewidth]{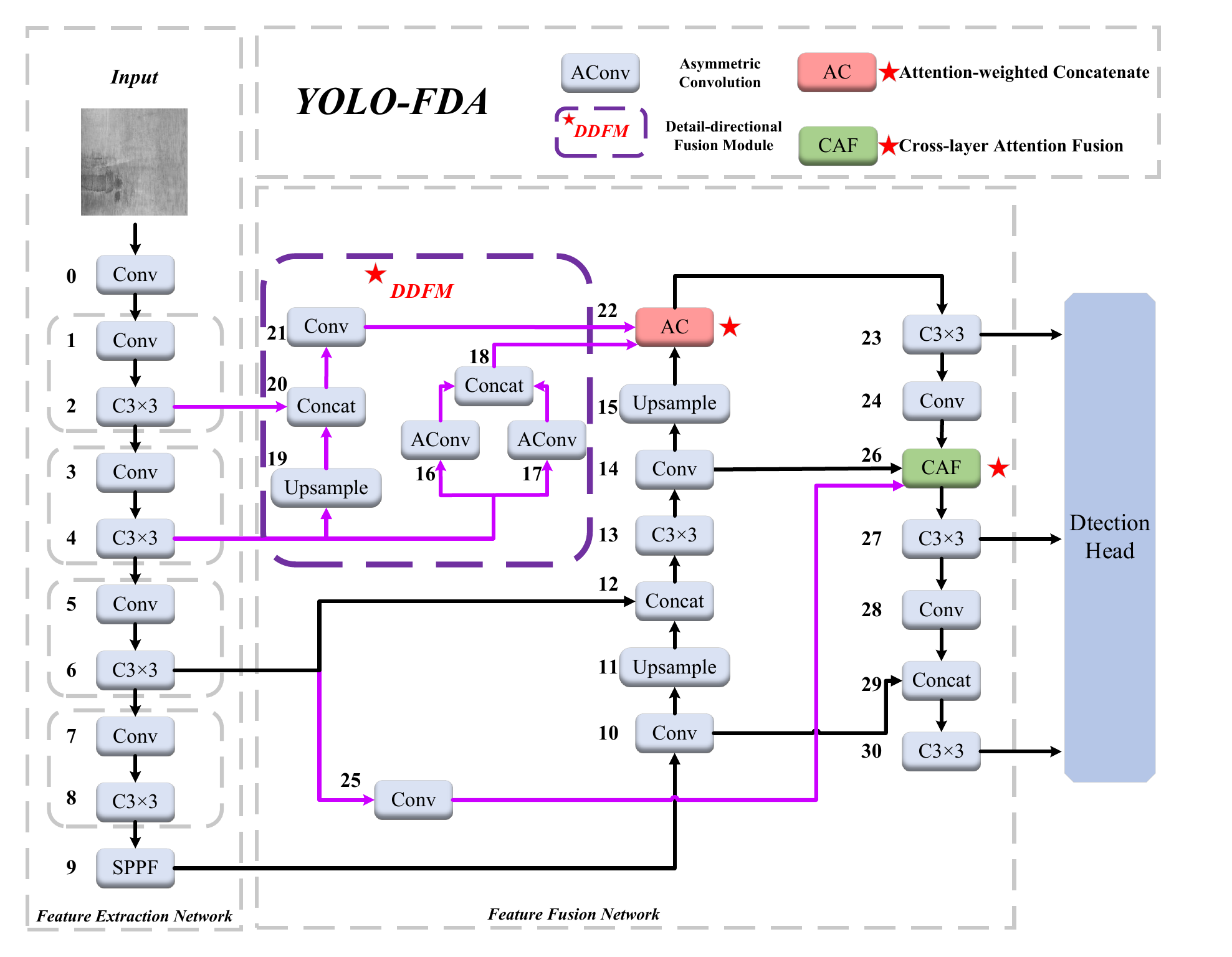}
\caption{Overall architecture of the proposed YOLO-FDA model.}
\vspace{-.8cm}
\label{fig:flow}
\end{figure*}

\begin{figure}[t]
    \centering
    \includegraphics[width=0.91\linewidth]{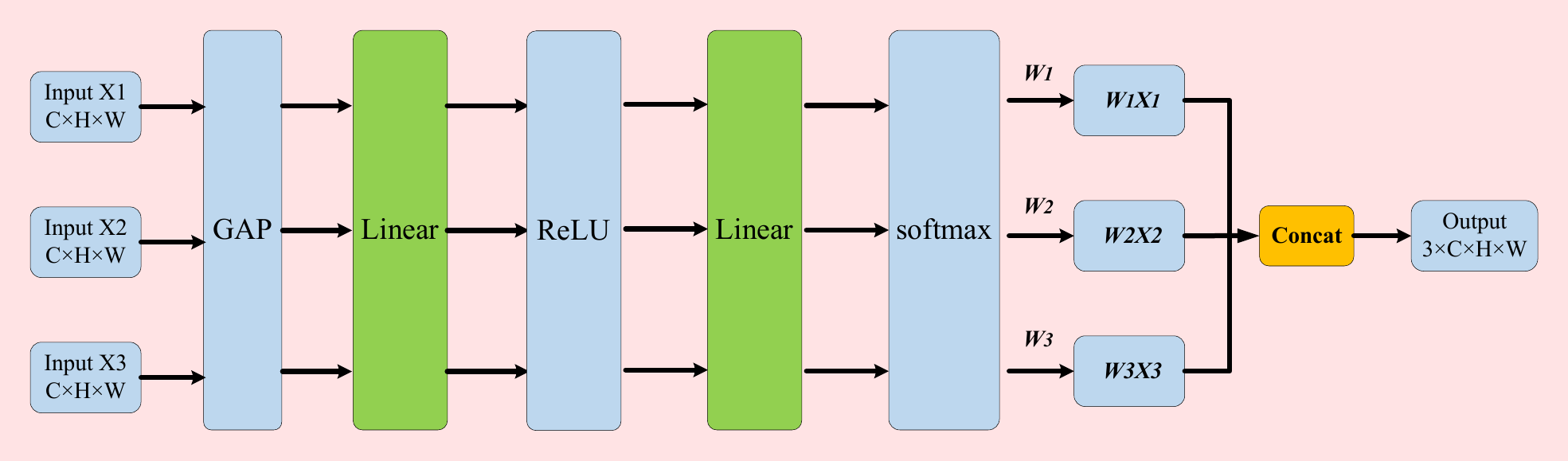}
\caption{Illustration of the proposed Attention-weighted Concatenation module.}
\vspace{-.4cm}
\label{fig:AC}
\end{figure}

\begin{figure}[t]
    \centering
    \includegraphics[width=0.91\linewidth]{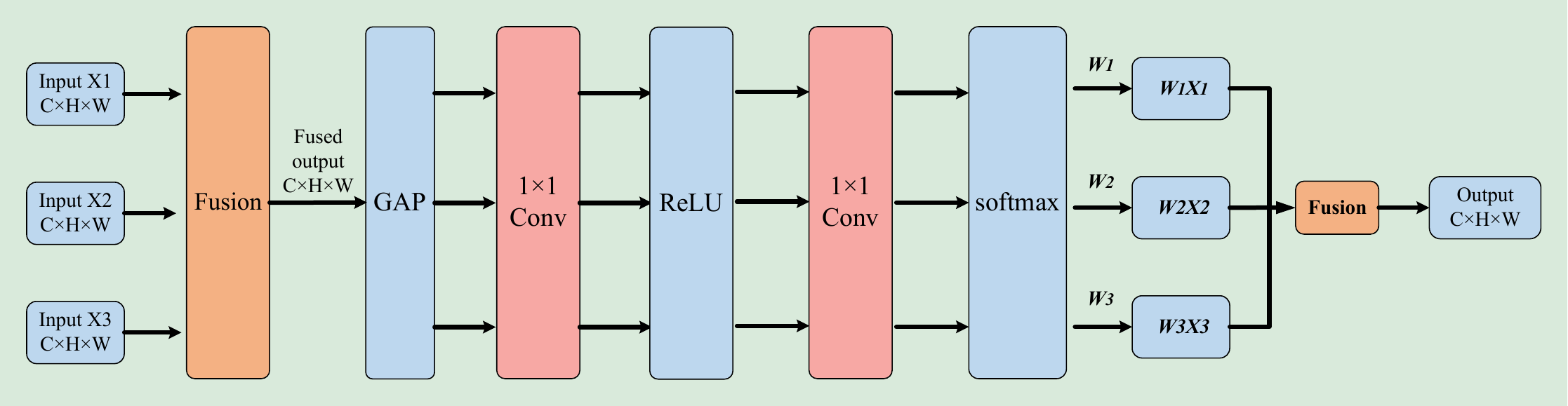}
\caption{Illustration of the proposed Cross-layer Attention Fusion module.}
\vspace{-.8cm}
\label{fig:CAF}
\end{figure}

The framework of our proposed YOLO-FDA is shown in Fig.\ref{fig:flow}. The basic YOLOv5 model consists of three parts: the backbone for feature extraction, the neck for feature fusion, and the head for generating bounding boxes, class probabilities, and objectness scores. Based on the baseline, we propose two innovations: (1) We add DDFM module on this basis to enhance the fusion of detail information; (2) AC and CAF modules are used at two position where multiple paths are fused to maintain information richness at the front-end of the network and reduce information redundancy and mutual interference in the later stage.

\subsection{Detail-directional Fusion Module}  
The basic YOLOv5 adopts the feature fusion method of PANet, which adds a bottom-up path on the basis of FPN, so that information can be transmitted bidirectionally and information fusion is strengthened. However, in the field of surface defect target detection, the requirements for detailed features and targets in different directions and with large aspect ratios are high, and the feature fusion of PANet and even BiFPN cannot meet the detection requirements. Therefore, we propose DDFM, integrating it into the feature fusion process.

As shown in the purple dotted box in Fig.\ref{fig:flow}, we propose DDFM, which needs to first upsample the output of the 4th layer of the YOLOv5 backbone to increase its length and width by 2 times, and then concatenate it with the output of the second layer, further enhancing the detailed information of the output feature map of the fourth layer in the basic YOLOv5 backbone, allowing the model to capture more detailed features. Additionally, the output feature map of the 4th layer of the backbone is subjected to two asymmetric convolutions and then concatenated. Specifically, in the asymmetric convolution process, we use $1 \times 7$ and $7 \times 1$ kernels to extract horizontal and vertical features respectively. Formally, these two operations can be expressed as:

Horizontal:
\vspace{-10pt}
\begin{equation}
F_h(x, y, c') = \sum_{k=-3}^{3} \sum_{c=1}^{C} w^h_{c', c, k} \cdot F(x, y + k, c),
\end{equation}\vspace{-20pt}

Vertical:
{
\setlength{\belowdisplayskip}{-20pt}%
\setlength{\abovedisplayskip}{-8pt}
\setlength{\abovedisplayshortskip}{-2pt}%
\setlength{\belowdisplayshortskip}{1pt}%
\begin{equation}
F_v(x, y, c') = \sum_{k=-3}^{3} \sum_{c=1}^{C} w^v_{c', c, k} \cdot F(x + k, y, c).
\end{equation}
}\noindent
Here, $F(x, y, c)$ denotes the input feature map at spatial location $(x, y)$ and channel $c$, while $F_h$ and $F_v$ represent the horizontally and vertically filtered output features, respectively. The asymmetric convolution weights are denoted as $w^h_{c', c, k}$ and $w^v_{c', c, k}$ for the horizontal and vertical directions.

To maintain the same number of channels as the original input, we set the output channel dimensions of each asymmetric convolution to half of the input channels. Specifically, if the input feature map has $C$ channels, both $F_h$ and $F_v$ will have $C/2$ channels. 
This direction-aware representation enhances the model's ability to distinguish defect types with strong directional features.
\subsection{Attention Fusion Module}

The feature fusion method of YOLOv5 is ordinary concatenation. Although the implementation cost is low and the amount of calculation is small, it will splice all features without selectivity and weight in the multi-path fusion process, resulting in information redundancy or conflict. Therefore, at the two three-path fusion nodes of YOLO-FDA, we consider using different attention fusion mechanisms: \textbf{AC} and \textbf{CAF}.

\noindent\textbf{AC.} In early fusion stages (layer 22), preserving detailed and diverse information is crucial. Therefore, we propose an AC module to weight each input feature map before concatenation, allowing the network to emphasize more informative sources while still retaining all input channels. The implementation process of the AC module is shown in Fig.\ref{fig:AC}.

Given a set of $N$ input feature maps $\{F_i\}_{i=1}^N$, each of shape $B \times C \times H \times W$, We first calculate their global average pooling results, preparing it for the subsequent fully connected attention network, which can be expressed as:
\vspace{-8pt}
\begin{equation}
g_i = \mathrm{GAP}(F_i) \in \mathbb{R}^{B \times C}.
\end{equation}\vspace{-20pt}

Each $g_i$ is passed through a two-layer fully connected attention network to generate scalar attention scores:
{
\setlength{\belowdisplayskip}{4pt}%
\setlength{\abovedisplayskip}{-2pt}
\begin{equation}
s_i = \mathrm{FC}(g_i) = W_2 \cdot \mathrm{ReLU}(W_1 \cdot g_i).
\end{equation}
}\noindent
Where $W_1 \in \mathbb{R}^{C' \times C}$ and $W_2 \in \mathbb{R}^{1 \times C'}$ are the weights of the two fully connected layers used to compute the attention score $s_i$, and $C' = C / r$ is the reduced dimensionality with reduction ratio $r$.

The normalized attention weights are obtained via softmax:
\vspace{-8pt}
\begin{equation}
\alpha_i = \frac{\mathrm{exp}(s_i)}{\sum_{j=1}^N \mathrm{exp}(s_j)}.
\end{equation}\vspace{-15pt}

Finally, the weighted feature maps are concatenated:
\vspace{-8pt}
\begin{equation}
F_{\text{AC}} = \mathrm{Concat}(\alpha_1 F_1, \alpha_2 F_2, \dots, \alpha_N F_N).
\end{equation}\vspace{-20pt}

This mechanism preserves the full channel information of all feature maps while emphasizing their relative importance. It is particularly effective in early fusion stages where feature redundancy is low but information diversity is high.

\noindent\textbf{CAF.} In deeper layers (layer 26), different features may contain overlapping or redundant information. To prevent overfitting and suppress repetitive detection of the same defect, we propose a CAF module that learns a weighted sum across layers rather than concatenation. The implementation process of the CAF module is shown in Fig.\ref{fig:CAF}.

Given $N$ input feature maps $\{F_i\}_{i=1}^N$, we first compute their average to obtain a representative feature map:
\vspace{-8pt}
\begin{equation}
F_{\text{avg}} = \frac{1}{N} \sum_{i=1}^{N} F_i.
\end{equation}\vspace{-10pt}

Next, we apply global average pooling to condense spatial information into a channel-wise descriptor:
\vspace{-4pt}
\begin{equation}
x = \mathrm{GAP}(F_{\mathrm{avg}}) 
     \in \mathbb{R}^{B \times C \times 1 \times 1}.
\end{equation}\vspace{-18pt}

To reduce parameters and extract a compact representation, a $1\times1$ convolution reduces the channel dimension from C to C'=C/r: 
{
\setlength{\belowdisplayskip}{-2pt}%
\setlength{\abovedisplayskip}{2pt}
\begin{equation}
s^{(1)} = W_1 * x \in \mathbb{R}^{B \times C' \times 1 \times 1}, \quad C'=\frac{C}{r}.
\end{equation}
}\noindent
Here, $W_1 \in \mathbb{R}^{C' \times C \times1 \times1}$ denotes the learnable weights of this convolutional layer.

We then apply ReLU activation to introduce non-linearity and promote feature sparsity:
\vspace{-5pt}
\begin{equation}
z = \mathrm{ReLU}(s^{(1)}) \in \mathbb{R}^{B \times C' \times 1 \times 1}.
\end{equation}\vspace{-18pt}

The reduced features are mapped to N channels, generating raw attention scores for each input map:
{
\setlength{\belowdisplayskip}{-2pt}%
\setlength{\abovedisplayskip}{0pt}
\setlength{\abovedisplayshortskip}{-2pt}%
\setlength{\belowdisplayshortskip}{0pt}%
\begin{equation}
s^{(2)} = W_2 * z \in \mathbb{R}^{B \times N \times 1 \times 1}.
\end{equation}
}\noindent
Here, $W_2 \in \mathbb{R}^{N \times C' \times1 \times1}$ represents the learnable weights of the second convolution.

To obtain comparable weights that sum to 1, we normalize the scores using Softmax over the N dimension:
\vspace{-12pt}
\begin{equation}
\boldsymbol{\beta} = \mathrm{Softmax}(s^{(2)}) \in \mathbb{R}^{B \times N \times 1 \times 1}.
\end{equation}\vspace{-20pt}

Finally, we apply the normalized weights \(\boldsymbol{\beta} = [\beta_1,\dots,\beta_N]\) to fuse the original feature maps:
\vspace{-8pt}
\begin{equation}
F_{\mathrm{CAF}}
= \sum_{i=1}^{N} \beta_i \, F_i,
\quad
F_{\mathrm{CAF}}\in\mathbb{R}^{B\times C\times H\times W}.
\end{equation}\vspace{-10pt}

This strategy effectively suppresses redundant responses and encourages the model to selectively attend to the most useful layer-wise features. Since the output has the same dimensionality as the inputs, it maintains structural consistency and computational efficiency.

\section{Experiment and Analysis}

To validate the proposed YOLO-FDA's superiority, it is compared with multiple defect detection approaches on two datasets, namely, GC10-DET and DAGM2007.

\subsection{Datasets}
\noindent\textbf{GC10-DET.} The GC10-DET ~\cite{1234gc10} dataset is a collection of 2294 grayscale images collected from real industrial environments, all of which are surface defects of various types of steel materials. The dataset classifies 10 different types of defects, namely punching (Pu), weld (Wl), crater (Cg), water spot (Ws), oil spot (Os), silk spot (Ss), inclusion (In), rolling pit (Rp), wrinkle (Crease) and waist fold (Wf), providing a wide range of resources for the development and testing of defect detection algorithms in the steel industry.

\noindent\textbf{DAGM2007.} Ten classes of artificially generated defects are shown in DAGM2007~\cite{4579745, dagm123}. The ground truth in the released dataset is an ellipse region containing a defect. We convert the ellipse region into a bounding box for defect detection using the DAGM2007 dataset. DAGM2007 contains 14,000 defect-free images and 2,100 defective images. It is a weakly supervised dataset and is saved in grayscale 8-bit PNG format.

\subsection{Evaluation Metrics}  
Following the criterion in~\cite{huang2024joining}, the Precision, Recall, and mAP are used to quantitatively evaluate the detection performance on the GC10-DET and DAGM2007, respectively.

\subsection{Implementation Details} 
\noindent\textbf{Training Details.} We implement our model on PyCharm with PyTorch 2.0.0 and train our network using an NVIDIA 4090D with 24 GB memory. According to the previous studies~\cite{9760388, huang2024joining}, we respectively divide the GC10-DET and DAGM2007 datasets into training, testing, and validation sets in a ratio of 8:1:1 by stratified random sampling. During the training phase, we set the input size as 640 $\times$ 640. For the details of the network architecture, our model is based on Yolov5 and uses CSPDarknet~\cite{wang2020cspnet} and SPPF~\cite{he2015spatial} as the backbone.

\noindent\textbf{Parameter Setup.} Furthermore, the model is optimized by the stochastic gradient descent (SGD) optimizer with a learning rate of 0.01, weight decay of 0.0005, and batch size of 8. The number of epochs is set to 250.

\subsection{Comparison with State-of-the-art Methods} 

\noindent\textbf{Comparisons on GC10-DET dataset.} We compare our proposed method with the representative defect detection methods on GC10-DET and DAGM2007. \autoref{tab: QUANTITATIVE RESULTS OF THE COMPARISON METHODS ON THE GC10-DET DATASET} shows our quantitative comparison with several representative defect detection methods on GC10-DET. The data of YOLOv3, YOLOv7\_tiny, YOLOv5n, YOLOV8n, YOLOv7, and DSL-YOLO are all from the literature~\cite{s24196268}, which is slightly different from our paper in terms of GPU type, etc., and its training number is 200 rounds. The data of YOLOv9s, YOLOv5s-improved are taken from paper~\cite{liSteelSurfaceDefect2025}, and it is also slightly different from this paper in terms of GPU type, etc., and the number of training rounds is 500, and the batch-size is 32. The results of YOLOv5s are obtained using the same experimental settings and environment as those described in this article. The YOLO-FDA model we propose achieves the highest mAP50 of 74.9\%  on GC10-DET, improving the baseline by 4.6\% and significantly surpassing many other YOLO series models, such as YOLOv7 and YOLOv9, which only achieve 69.4\% and 70.2\% respectively in the mAP50 index. Secondly, the YOLO-FDA also performs well in the precision index, reaching 76.9\%, outperforming other representative YOLO models. In terms of recall metric, our model performs well as well, achieving a recall of 71.7\%, which is higher than other classic models.
\begin{table*}[t]
\centering
\caption{Quantitative results of the comparison methods on the GC10‑DET dataset.}
\label{tab: QUANTITATIVE RESULTS OF THE COMPARISON METHODS ON THE GC10-DET DATASET}
\resizebox{0.6\columnwidth}{!}{%
\begin{tabular}{ccccc}
\hline
Model                                  & Precision (\%) & Recall (\%) & mAP50 (\%) \\ \hline
\multicolumn{1}{c|}{YOLOv3~\cite{redmon2018yolov3}}               & 68.9        & 52.2     & 61.8     \\
\multicolumn{1}{c|}{YOLOv7\_tiny~\cite{wang2023yolov7}}         & 59.0        & 63.8     & 62.6     \\
\multicolumn{1}{c|}{YOLOv5n~\cite{song2021object}}             & 64.9        & 68.1     & 68.3     \\
\multicolumn{1}{c|}{YOLOv8n~\cite{lv2024steel}}              & 61.9        & 66.5     & 68.3     \\
\multicolumn{1}{c|}{YOLOv7~\cite{wang2023yolov7}}               & 65.3        & 62.1     & 69.4     \\
\multicolumn{1}{c|}{YOLOv9s~\cite{liSteelSurfaceDefect2025}}             & 68.2        & 67.7     & 70.2     \\
\multicolumn{1}{c|}{YOLOv5s~\cite{song2021object}}              & 70.8        & 69.4     & 70.3     \\
\multicolumn{1}{c|}{YOLOv5s-improved~\cite{liSteelSurfaceDefect2025}}   & 71.9        & 68.4     & 71.0     \\
\multicolumn{1}{c|}{DSL-YOLO~\cite{s24196268}}             & 70.5        & 67.8     & 72.5     \\
\multicolumn{1}{c|}{Ours}                 & \textbf{76.9}        & \textbf{71.1}      & \textbf{74.9}          \\ \hline
\end{tabular}%
}
\vspace*{-0.7cm}
\end{table*}

\noindent\textbf{Comparisons on DAGM2007 dataset.} DAGM2007 is an artificially generated image and contains ten different background texture patterns, and there are more small targets. \autoref{tab: Comparisons on DAGM2007 DATASET} shows the training results of our proposed model and some representative algorithms on DAGM2007. Among them, the data of SSD, Faster RCNN, RetinaNet and Cascade R-CNN are all from the literature~\cite{xu2021msb}, and the environment and parameters used are slightly different from those of this paper, such as the number of training epochs and learning rate. The data of YOLOv5n, YOLOv5s, HIC-YOLOv5, YOLOv7\_tiny and YOLOv9 are trained using the same environment and parameters as this paper. As can be seen from the table, the YOLO-FDA model proposed by us performs well in many indicators. First, notably, our model achieves 67.9\% in the mAP50-95 indicator, which improves the effect of YOLOv5 by 2.7\% and exceeds the other models in the table. In terms of mAP comparison of various categories, our model also performs outstandingly when testing C1, reaching 84.5\%, far exceeding many classic models. At the same time, in the detection tasks of C2, C4, C5, C6, C8, and C9, our model performance is also at the forefront, reaching 76.7\%, 72.7\%, 77.1\%, 74.4\%, 52.6\%, and 70.5\% respectively. The above data all reflect the excellent performance of YOLO-FDA.

\begin{table*}[t]
\centering
\caption{Quantitative results of the comparison methods on the DAGM2007 dataset.}
\label{tab: Comparisons on DAGM2007 DATASET}
\resizebox{1\textwidth}{!}{%
\begin{tabular}{cccc|cccccccccc}
\hline
\multirow{2}{*}{Methods}            & \multicolumn{1}{l}{\multirow{2}{*}{mAP50-95 (\%)}} & \multicolumn{1}{l}{\multirow{2}{*}{Precision (\%)}} & \multicolumn{1}{l|}{\multirow{2}{*}{Recall (\%)}} & \multicolumn{10}{c}{mAP value of each catagory}                                                                                                                                                                                                                                                  \\ \cline{5-14} 
                                    & \multicolumn{1}{l}{}                                                 & \multicolumn{1}{l}{}                               & \multicolumn{1}{l|}{}                            & \multicolumn{1}{l}{C1 (\%)} & \multicolumn{1}{l}{C2 (\%)} & \multicolumn{1}{l}{C3 (\%)} & \multicolumn{1}{l}{C4 (\%)} & \multicolumn{1}{l}{C5 (\%)} & \multicolumn{1}{l}{C6 (\%)} & \multicolumn{1}{l}{C7 (\%)} & \multicolumn{1}{l}{C8 (\%)} & \multicolumn{1}{l}{C9 (\%)} & \multicolumn{1}{l}{C10 (\%)} \\ \hline

\multicolumn{1}{c|}{YOLOv7\_tiny~\cite{wang2023yolov7}}       & 60.5                                                                & 88.3                                              & 95.2                                            & 67.0                       & 71.0                       & \textbf{65.4}                       & \textbf{74.8}                       & 67.9                       & 32.3                       & 46.9                       & 40.8                       & 67.1                       & 62.6                        \\
\multicolumn{1}{c|}{SSD~\cite{liu2016ssd}}            & 63.2                                                                 & \textbf{96.0}                                              & 96.4                                            & 52.6                       & 67.3                       & 56.1                       & 68.5                       & 71.7                       & 65.8                       & 59.5                       & 48.2                       & 69.7                       & 72.6                        \\
\multicolumn{1}{c|}{HIC-YOLOv5~\cite{10610273}} & 64.2                                                                 & 87.7                                                 & 94.3                                               & 73.5                       & 72.9                       & 61.6                       & 64.2                       & 75.0                       & 50.5                       & 56.3                       & 45.4                       & 70.6                       & 71.7                        \\
\multicolumn{1}{c|}{Faster RCNN~\cite{Shaoqing}}    & 64.6                                                                 & 75.2                                              & \textbf{98.6}                                            & 60.5                       & 64.4                       & 56.5                       & 69.2                       & 69.6                       & 70.6                       & 58.8                       & 51.9                       & \textbf{72.2}                       & 72.0                        \\
\multicolumn{1}{c|}{YOLOv5n~\cite{1234yolov5}}       & 64.9                                                                 & 95.6                                              & 98.0                                            & 71.3                       & 75.7                       & 43.6                       & 74.5                       & 74.8                       & 62.8                       & 60.8                       & 45.4                       & 68.8                       & 71.2                        \\
\multicolumn{1}{c|}{RetinaNet~\cite{ouyang2017chained}}      & 65.2                                                                 & --                                                 & --                                               & 62.5                       & 66.9                       & 56.0                       & 70.2                       & 72.8                       & 71.7                       & 60.9                       & 53.7                       & 69.8                       & 71.9                        \\
\multicolumn{1}{c|}{YOLOv5s~\cite{1234yolov5}}       & 65.2                                                                 & 91.1                                              & 98.0                                            & 75.9                       & 66.5                       & 61.4                       & \textbf{74.8}                       & 72.9                       & 57.6                       & 52.7                       & 49.6                       & 69.6                       & 71.3                        \\
\multicolumn{1}{c|}{Cascade  R-CNN~\cite{cai2018cascade}} & 66.0                                                                 & --                                                 & --                                               & 63.1                       & 66.2                       & 56.4                       & 66.5                       & 72.0                       & \textbf{75.4}                       & 62.8                       & \textbf{53.0}                       & 69.8                       & \textbf{73.4}                        \\
\multicolumn{1}{c|}{YOLOv9~\cite{wang2024yolov9}} & 67.1                                                                 & 95.6                                                 & 93.8                                               & 61.6                       & \textbf{78.7}                       & 60.5                       & 71.0                       & \textbf{78.7}                       & 63.3                       & \textbf{64.8}                       & 47.6                       & 71.2                       & 71.7                        \\
\multicolumn{1}{c|}{Ours}           & \textbf{67.9}                                                                 & 94.7                                              & 98.4                                            & \textbf{84.5}                       & 76.7                       & 52.9                       & 72.7                       & 77.1                       & 74.4                       & 48.3                       & 52.6                       & 70.5                       & 69.6                        \\ \hline
\end{tabular}%
}
\vspace*{-0.75cm}
\end{table*}

\subsection{Ablation Studies and Analysis}  
\begin{figure}[t]
    \centering
    \includegraphics[width=0.9\linewidth]{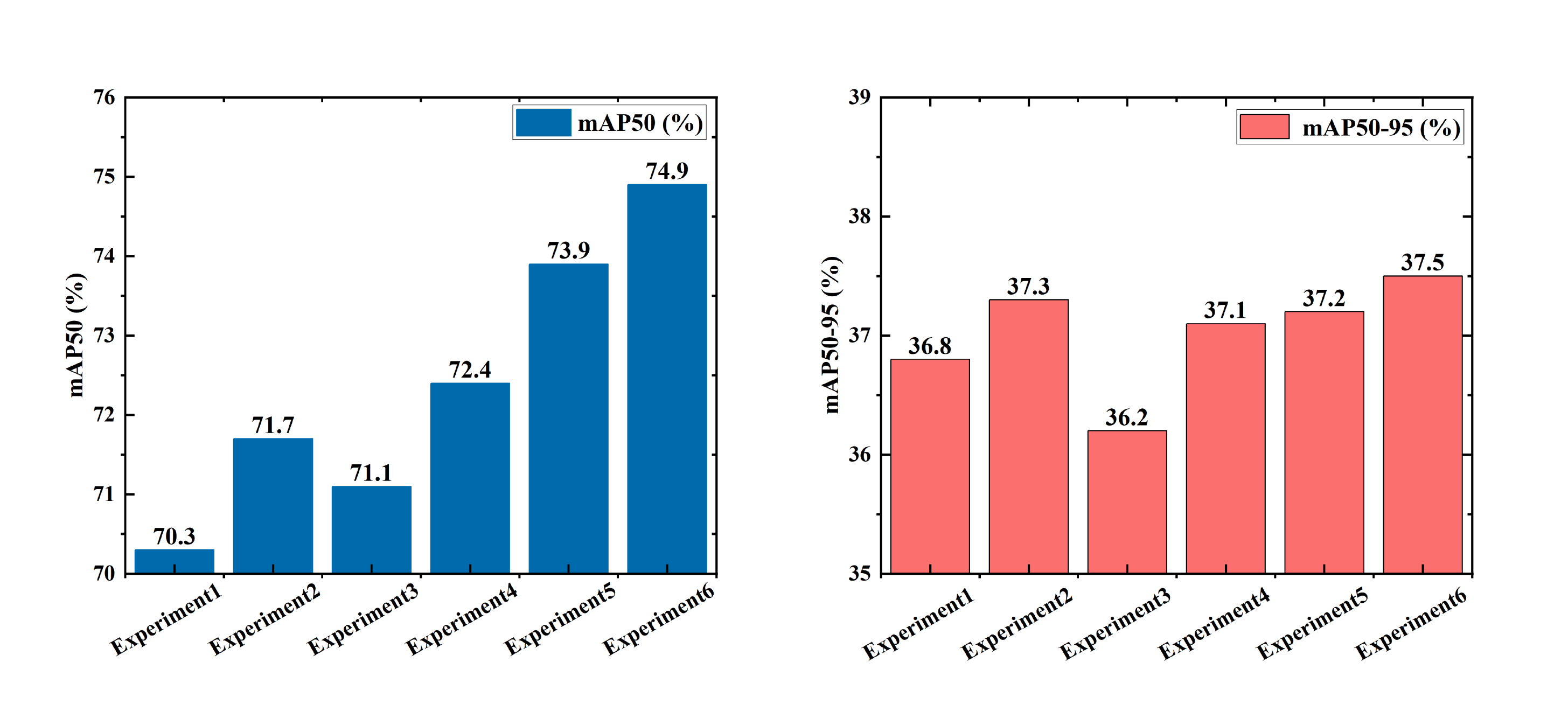}
\caption{Quantitative results of different ablation models tested on the GC10‑DET dataset.}
\vspace{-.8cm}
\label{fig:ablation}
\end{figure}

In order to demonstrate the role and significance of each innovative point of the proposed model, we design six ablation experiments and conduct experiments on GC10-DET. The results are shown in the Fig.\ref{fig:ablation}. The experimental environment and parameter settings of all experiments are consistent with those described previously in this paper. The first experiment is conducted only on YOLOv5, which is used for comparison with other ablation models in the following; the second experiment replaces the neck part of YOLOv5 from PANet to BiFPN to verify the effect of the BiFPN architecture; the third experiment adds DDFM on the basis of BiFPN to show the performance changes brought by this innovation; the fourth experiment adds CAF module on the basis of the third experiment to reflect the effect of it; in order to reflect the different effects of AC and CAF modules in the feature fusion process, we design the fifth experiment, adding the AC module instead of CAF to the basis of the third experiment; the sixth experiment combines all the innovations proposed in this paper to reflect the effect of the collaborative work of each module.


\begin{figure}[t]
    \centering
    \includegraphics[width=0.85\linewidth]{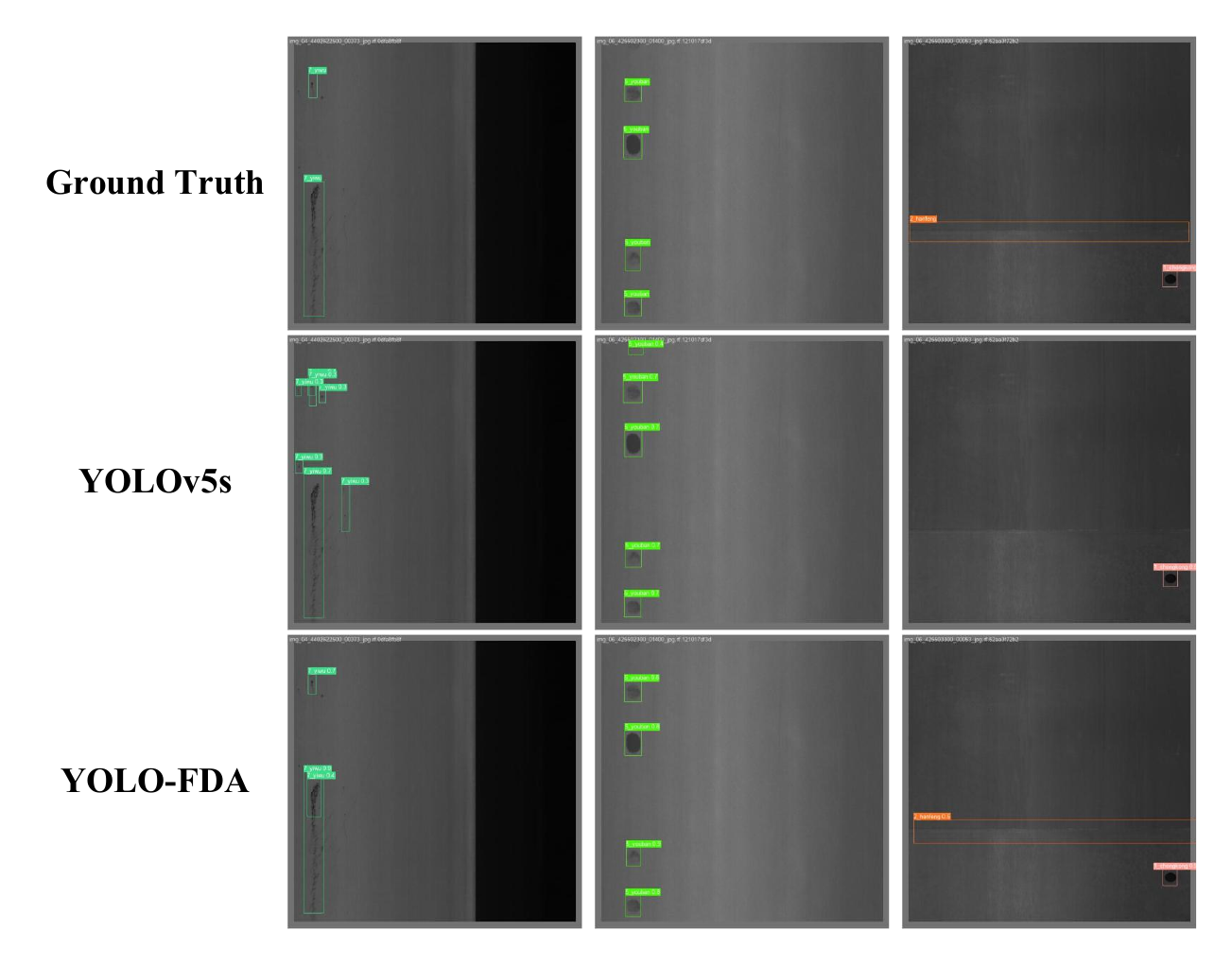}
\caption{Visualization results. Compared our method with baseline. The cyan frame is a defect called inclusion, the green frame is a defect called OilSpot, the orange frame is a defect called OilSpot, and the pink frame is a defect called Punching.}
\vspace{-.87cm}
\label{fig:comparison}
\end{figure}

In what follows, the proposed YOLO-FDA is comprehensively analyzed from four aspects to investigate the logic behind its superiority.

\noindent\textbf{Role of BiFPN.} From Experiment 2, we can see that after using the BiFPN architecture in the feature fusion part of YOLOv5, mAP50 and mAP50-95 indicators are improved, 1.4\% and 0.5\% higher than the baseline, respectively. This is mainly because BiPFN adds additional residual connections between levels of the same scale, making the propagation of features in the horizontal level more sufficient.

\noindent\textbf{Influence of DDFM.} From Experiment 3, we can see that after adding the DDFM module on the basis of BiFPN, the mAP50 and mAP50-95 indicators have slightly decreased. By analyzing the detection results, we believe that this is because this improvement enhances the sensitivity to small target details and defects in different directions, but still uses a simple concatenate method when fusion, which will cause a lot of information to be redundant, affecting the final judgment of the detection head and reducing the mAP indicators.

\noindent\textbf{Impact of AC and CAF.} From Experiments 4 and 5, we can see that different feature fusion methods of the AC module and the CAF module will bring different effects. If only the CAF module is used, the training results of YOLOv5 can be improved by 2.1\% and 0.3\% in mAP50 and mAP50-95 respectively; while if only AC is used, the results can be improved by 3.6\% and 0.4\% respectively. We believe that this is because although the AC fusion method can learn the weights of different feature maps through attention, it still uses the concatenate method in the end, which may still cause inappropriate information fusion. Fusion of features using only the CAF method will cause information loss in the early stage of fusion, resulting in feature saturation in the model learning in the later stage of training and reaching an information bottleneck.

\noindent\textbf{Overall effect.} Experiment 6 shows that the mixed use of AC and CAF modules in the feature fusion stage can bring better results than using only one of them. Combined with the DDFM, the YOLO-FDA model we proposed can achieve 74.9\% in mAP50 and 37.5\% in mAP50-95 on GC10-DET, which are 4.6\% and 0.7\% higher than the baseline respectively.

\subsection{Visualization}  

As can be seen from Fig.\ref{fig:comparison}, the YOLO-FDA model we proposed can more comprehensively detect some defects with larger aspect ratios or small target defects due to the addition of a fusion path for detail features and the increased sensitivity to multi-scale and multi-directional features. Meanwhile,  compared with the YOLOv5 model, our model is more conducive to reducing the phenomenon of repeated detection.

\section{Conclusion}\label{sec:con} 
In this work, we present YOLO-FDA, a fine-grained and attention-enhanced YOLO-based framework for surface defect detection. By integrating bidirectional feature fusion, directional detail enhancement, and novel attention-based modules, YOLO-FDA effectively improves multiscale robustness and defect sensitivity. Experimental results demonstrate superior performance over existing methods across various defect types and scales, highlighting the model’s effectiveness and generalizability in industrial inspection scenarios.
%
%
%
%

{\small
\bibliographystyle{IEEEtran}
\bibliography{ref}
}

\end{document}